\documentclass[a6paper, 10pt, conference]{ieeeconf} 
\usepackage{fancyhdr}
\usepackage{graphicx}
\usepackage{float}
\usepackage[bottom]{footmisc}
\usepackage{amsmath}
\usepackage{enumerate}
\usepackage{xcolor}

\IEEEoverridecommandlockouts  

\title{\LARGE \bf
Evaluating the Effectiveness of Automated Identity Masking (AIM) Methods with Human Perception and a Deep Convolutional Neural Network (CNN)}

\author{\parbox{16cm}{\centering
    {\large  {Kimberley D. Orsten-Hooge\textsuperscript{$*$}\thanks{*K.O.H and A.B. contributed equally to this work.}}$^1$, Asal Baragchizadeh\textsuperscript{$*$}$^1$,  Thomas P. Karnowski$^2$, David S. Bolme$^2$, Regina Ferrell$^2$, Parisa R. Jesudasen$^1$, Carlos D. Castillo$^3$,
    and Alice  J.  O'Toole$^1$ }\\
    {\normalsize
    $^1$ School of Behavioral and Brain Sciences, The University of Texas at Dallas, Richardson, TX, USA\\
    $^2$ Oak Ridge National Laboratory, Oak Ridge, TN, USA\\
    $^3$ University of Maryland, USA\\}}}

\begin{document}
\maketitle
\begin{abstract}
Face de-identification algorithms have been developed in response to the prevalent use of public video recordings and surveillance cameras. Here, we evaluated the success of identity masking in the context of monitoring drivers as they actively operate a motor vehicle. We studied the effectiveness of eight de-identification algorithms using human perceivers and a state-of-the-art deep convolutional neural network (CNN). 
We used a standard face recognition experiment in which human subjects studied high-resolution (studio-style) images to learn driver identities. Subjects were tested subsequently on their ability to recognize those identities in low-resolution videos depicting the drivers operating a motor vehicle. The videos were in either unmasked format, or were masked by one of the eight de-identification algorithms. 
All masking algorithms lowered identification accuracy substantially, relative to the unmasked video. In all cases, identifications were made with stringent decision criteria indicating the subjects had low confidence in their decisions. 
We also examined the performance of the CNN on identification. First, we tested identity matching between the high-resolution images and templates created from both the masked and unmasked videos.
Next, we tested CNN on matching identities between the high-resolution images and masked videos, and between the unmasked and masked videos. CNN performance on both tasks was comparable to human behavior, including yielding different results across the mask types.
The results for both human and CNN indicated that even simple methods such as edge-detection masking and color inversion can greatly alter identification performance. We propose that carefully tested de-identification approaches, used alone or in combination, can be an effective tool for protecting the privacy of individuals captured in videos. We note that no approach is equally effective in masking all stimuli, and that future work should examine possible methods for determining the most effective mask per individual stimulus.
 \end{abstract}

\section*{INTRODUCTION}
Video recording technologies such as head-mounted devices (e.g., Google Glass), cloud-based video surveillance, body and dashboard cameras used by law enforcement agencies, and civilian drones have emerged in both public and private arenas. These technologies are often used without the explicit consent or even awareness of the person being recorded. It is, therefore, not surprising that the widespread use of video surveillance has raised privacy concerns. As such, there has been great interest in the development of de-identification algorithms in computer vision. These algorithms typically have two objectives: 1) to effectively mask the identity of individuals being recorded, thereby preserving their privacy, and 2) to retain the important information for which the recording was made.

Various algorithmic techniques have been employed in the de-identification of still frontal face images in optimal lighting conditions (e.g., ad-hoc distortion \cite{Boyle2000, Hudson1996} and {\it k}-same \cite{Newton2003}). However, these techniques are limited in their ability to track and mask faces in dynamic video stimuli \cite{Ribaric2015}. Recent work involving dynamic video stimuli has employed approaches that cover motion detection without the need for video decryption \cite{Ma2018}, data hiding \cite{Cheung2009}, motion detection followed by obfuscation \cite{Wang2018,Li2017}, foveation \cite{Alonso2017}, singular value decomposition \cite{Chriskos2017}, or full face/body de-identification using segment replacement \cite{Brkic2017}. 

The continued challenge of these approaches is to maintain the integrity of the data being masked (e.g., expression/emotion/action preservation), while eliminating discriminating cues about specific identity. As such, the tradeoff between data preservation and de-identification has limited the effectiveness of dynamic facial masking techniques. Moreover, although high facial de-identification rates have been achieved, those rates never reach 100\%.  

\section*{Goals}
In the present study, we address the question of de-identification without consideration of the problem of data preservation. We do this under the assumption that the latter cannot be reasonably addressed until the former problem is solved. This study is aimed at evaluating the effectiveness of eight different identity masking algorithms, which we apply to the Second Strategic Highway Research Program's Naturalistic Driving Study (SHRP2-NDS) video dataset \cite{Campbell2012}. Notably, we use the human perceptual system as a standard of de-identification success, as well as a state-of-the-art deep convolutional neural network (CNN). Specifically, we ask whether a human and our selected CNN can recognize a face that has been masked with a de-identification algorithm.  

\section*{Background}
\subsection{SHRP2-Naturalistic Driving Study Data}
The  identity masking challenges posed by the SHRP2-NDS dataset are nearly unique in their scale as well as in the range of imaging conditions encompassed in the data. The SHRP2-NDS dataset includes approximately 2 petabytes of video footage from approximately $3,400$ drivers obtained over 1 to 2 years of observation. However, the dynamic video nature of the dataset provides for highly salient, personally identifiable information about each of the drivers. This information must be eliminated to allow for wider distribution, examination, and analysis of the driving data by various institutions with research interests in said data. In particular, transportation researchers who have signed data usage licenses would benefit greatly from access to masked facial data, advancing the utility of the SHRP2-NDS to answer important scientific questions about driver behavior.

Although much progress has been made in the field of identity-masking, the naturalistic nature of the SHRP2-NDS dataset poses difficult challenges to available de-identification algorithms. The dataset is characterized by extreme illumination conditions (e.g., night-time shadowing, day-time bright spots, or illumination via transient headlights as a car turns). There is also the problem of quick driver movements (e.g., head turns and other actions which are very common in real-world driving). Despite these challenges, the research opportunities afforded by a dataset of this quality are numerous and diverse. The need to make this dataset available to researchers has motivated these de-identification efforts and associated evaluations to assure masking success.

\subsection{Identification using human perception}
This work is a follow-up to a previous study \cite{Baragchizadeh2017} in which the effectiveness of both identity disruption and action preservation were measured using human evaluators. In that study \cite{Baragchizadeh2017}, de-identification success was examined for driver videos taken from a research-available test set (the Virginia Tech Transportation Institute (VTTI) Head Pose Validation, HPV dataset) modeled on the SHRP2-NDS data. Specifically, in \cite{Baragchizadeh2017}, the test was aimed at determining whether human subjects could recognize drivers whose faces were masked with one of two algorithms. The first algorithm was the ``personalized supervised bilinear regression method for Facial Action Transfer (FAT)" \cite{Huang2012}. The second algorithm was an alternative edge-masking method used to ``fill-in" video frames in which the FAT mask failed to adequately track and mask a driver's face. Human performance accuracy was measured by the signal detection measure of {\it d'}.

The human experimental data showed that both of the algorithms substantially reduced the accuracy of human identification of drivers \cite{Baragchizadeh2017}. These findings are generally consistent with other studies that have used non-human evaluation methods of de-identification and masking algorithms other than the FAT and edge-masking methods used by \cite{Baragchizadeh2017} (cf., \cite{Meden2017, Chen2018, Mirjalili2018}).

Notably, the human performance data in \cite{Baragchizadeh2017} also indicated that both of the masking algorithms greatly lowered human confidence in decisions about recognition (as measured by {\it C}, a signal detection measure of response bias). Specifically, both masking algorithms increased the likelihood that subjects would respond that they ``did not recognize the driver''. Thus, although neither algorithm was completely successful at eliminating the possibility of face identification by human perceivers, recognition judgments revealed a high degree of uncertainty. This is an important result as it suggests that even successful identifications were made with very low confidence.

The previous work provides a starting point for evaluating identity masking success for the SHRP2-NDS dataset, but leaves a number of important questions open for further exploration. First, in \cite{Baragchizadeh2017}, only two algorithms were evaluated. Here, we consider six additional masking methods. This larger and more diverse set of algorithms offers a more comprehensive evaluation of possible approaches to de-identification of this challenging set of videos. Second, because we focused here only on the identity masking question,
we were able to use videos that limited abrupt driver motions (e.g., using a cell phone, checking rearview mirror, head turn, etc). This eliminated the problems the algorithms previously had with locating and tracking faces, thereby assuring that the mask was correctly placed over the drivers' faces at all times. Lastly, in \cite{Baragchizadeh2017}, the ``studio style" training images shown to subjects included peripheral cues like hair and ears. Here, we eliminate these artifactual cues to assure that identification is based only on the internal face.

\subsection{Identification using CNN}
For more than two decades, computer-based facial recognition systems have been used by
forensic face examiners (e.g., passport controls and immigration visas) \cite{White2015}, commercial technology (e.g., facial recognition based security features on smartphones), advertising, and in healthcare applications.
Prior to 2013, facial recognition algorithms were modeled on a variety of methods, such as Principal Component Analysis (PCA), Linear Discriminant Analysis (LDA), eigenfaces, and support vector machines \cite{Zhao1998, Turk1991, Blanz2003}. In the best cases, the performance of these algorithms matched humans (cf. \cite{Bruce1999, Burton2010, Kemp1997}), but only when the tests were done with highly constrained, high-quality, studio-style, frontal images of faces with limited changes in pose, expression, and illumination \cite{Phillips2018, OToole2008, Phillips:2007fk, OToole:2007fk, Phillips98b}. 

Since then, algorithms based on CNNs have shown impressive gains in face recognition accuracy on problems that require generalization across substantial changes in image characteristics. Inspired by the feed-forward hierarchical organization of the visual cortical areas of primates, CNNs consist of cascaded layers of neuron-like units that convolve and pool the data across the layers. CNNs are trained typically on millions of images; for face recognition applications, these images consist of thousands of face identities. The training data often consist of unconstrained, low quality images with substantial variations in pose, illumination, and expression \cite{Phillips2018, Parkhi2015, Ranjan2018, Sankaranarayanan2016, Sun2015}. The output is a compact numerical feature vector that represents a face. This face descriptor supports highly robust, accurate face recognition.

In this study, as a comparison for humans, we tested whether a high-performing CNN (\cite{bansal2017s, ranjan2018crystal}) could recognize faces in the masked videos. The network we tested achieved state-of-the-art results for both unconstrained face verification and face recognition on the challenging IARPA JANUS Benchmark-A (IJB-A) \cite{Klare2015}, IARPA-JANUS Benchmark-B (IJB-B) \cite{Whitelam2017}, IARPA-JANUS Benchmark-C (IJB-C) \cite{Maze2018}, and Labeled Face in the Wild (LFW) \cite{Huang2008} datasets.

The contributions of the present work are as follows:

\begin{itemize}
\item We provide the first comparison of multiple masking methods for face de-identification, using humans as the standard of identification accuracy, as well as a state-of-the-art CNN;
\item We control for identification based on factors other than the face (e.g., hair style, ear shape, etc.);
\item We control for de-identification failures that might have occurred due to inaccurate applications of the mask to the face;
\item We offer empirical validation of the use of the human perceiver as an evaluator for the success of de-identification; and
\item We provide a useful replication of the results of a previous study suggesting that single masking approaches might be combined to more effectively mask faces of different individuals. 
\end{itemize}

\section{EXPERIMENT I: Evaluating the Effectiveness of Automated Identity Masking (AIM) Methods with Human Perception}

In this experiment, we examined the success of various algorithms at masking the identity of drivers in low-resolution videos as they actively operated a motor vehicle. Human observers were used to evaluate the performance of each algorithm. 
 
\subsection{Methods}
\subsubsection{Dataset}

As in \cite{Baragchizadeh2017}, we used the VTTI-generated Head Pose Validation (HPV) dataset of driver videos and still images for this test. This dataset was developed for research uses and does not have the same degree of privacy restrictions as the SHRP2-NDS video dataset. The HPV dataset was acquired using the same equipment as the SHRP2-NDS videos, and therefore achieves comparable resolution, compression rates, and illumination \cite{Paone2015}. By closely simulating the quality of the SHRP2-NDS data, the masking methods applied here are assumed to be applicable to the full range of SHRP2-NDS stimuli. Moreover, the data we report should generalize well to the full SHRP2-NDS dataset.
 
The HPV dataset includes 41 high-resolution, front-facing, color, still JPEG images of driver faces. The images were tightly cropped around the face and placed on a black background. Examples from a similar dataset are shown in Figure 1. This cropping eliminated features such as ears, hair, and some aspects of face shape as identification cues. The original still image resolution for all drivers was $4320\times3240$ pixels. However, because the stimuli were cropped, variations between individual drivers' face width and height resulted in image sizes ranging from 1331 to 1984 pixels in width and 1324 to 2080 pixels in height. Note, however, that standardizing the width/height of the cropped images would have altered the quality of the images. Therefore, we left the original aspect ratios intact. 

    \begin{figure}[h]
      \centering
      \includegraphics[width=.47\textwidth]{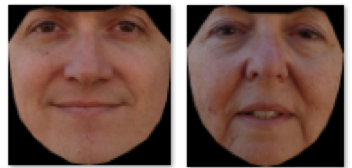}
      \caption{Examples of how the images used for training were cropped. For confidentiality purposes, these are not actual stimuli used in the experiment}
      \label{figurelabel}
   \end{figure}

The HPV dataset also includes low-resolution ($384\times259$ pixels, 14.99FPS) videos of each driver actively driving a car or performing staged activities commonly executed while driving. Videos were shown either unmasked (face of the driver was not altered) or masked with one of 8 masking algorithms (description of algorithms follows). 

Each video segment was 5s long. The segments for each driver were identical between mask conditions to allow for more direct comparison of mask effectiveness between conditions. Because preservation of driving actions under different mask conditions (e.g., using a cell phone, checking rearview mirror, etc.) was not a critical research question in the current study, driver motions and actions were limited in these videos. This reduced movement assured that the drivers' faces were masked at all times. It also prevented the mask algorithms 
from ``losing" the drivers' faces as a result of too much motion. As such, unsuccesful identity masking could not be attributed to the incorrect application of the mask.   

\subsubsection{Automated Identity Masking Algorithms}
In addition to the FAT masking method used in \cite{Baragchizadeh2017} and described above,  we also examined the following  masking conditions (Fig. 2). 

\begin{itemize}
\item The {\bf DMask method} \cite{DMask} renders a generic avatar face over any face detected in the video. This is achieved by tracking and extracting eye gaze, face position, head position, eye and mouth movement, and facial features, in order to synthesize appropriate facial motion. 
\item The {\bf Canny method} \cite{Canny1986} performs a set of operations designed to produce optimal edge detection, including applications of a Gaussian smoothing filter, a set of gradient-based edge detectors to enhance edges in the image, then a non-maximum suppression, threshold, and tracking to produce thin, refined edges. 
\item The {\bf Scharr method} \cite{Jahne1999} performs a more fundamental edge detection based on the Scharr operator, which filters the image to enhance edges and approximate the image gradient. 
\item A combinatorial approach dubbed here the {\bf OverMask method} was employed by overlaying the Canny and FAT methods. 
\item In addition, we used inverted color versions of the Canny, Scharr, and OverMask methods.  
\end{itemize}

Note that the FAT, OverMask, and DMASK methods required a face and landmark detection phase, which was not needed in the Canny and Scharr-based methods. As shown in Fig. 2 (Mask conditions), the masking methods result in different image outputs with potential differences in physical perception for recognition. The implemented methods also differ in processing time.
Table 1 shows processing time using the default implementations with an Intel Xeon CPU E5-1620 v3\textsuperscript{\textregistered} 3.5GHZ with 64 GB RAM running the Windows 10 operating system.

\begin{table}[h]
\caption{De-identification Processing Time by Mask Algorithm}
\label{table_example}
\begin{center}
\scalebox{1.2}{
\begin{tabular}{|l||c||l|}
\hline
 Mask Condition & Processing Rate & Time (s)\footnotemark\\
\hline
Scharr & 30.06 & 1.2\\
\hline
Scharr-inverted & 29.45 & 2.037\\
\hline
OverMask & 2.27 & 26.49\\
\hline
OverMask-inverted & 2.24 & 26.81\\
\hline
DMask & 0.29 & 208.68\\
\hline
FAT & 3.3 & 18.21\\
\hline
Canny & 25.77 & 2.33\\
\hline
Canny-inverted & 25.32 & 2.37\\
\hline
\end{tabular}}
\end{center}
\end{table}
\footnotetext{Time in seconds to process 60s of video data.}

   \begin{figure}[h]
      \centering
      \includegraphics[width=.5\textwidth]{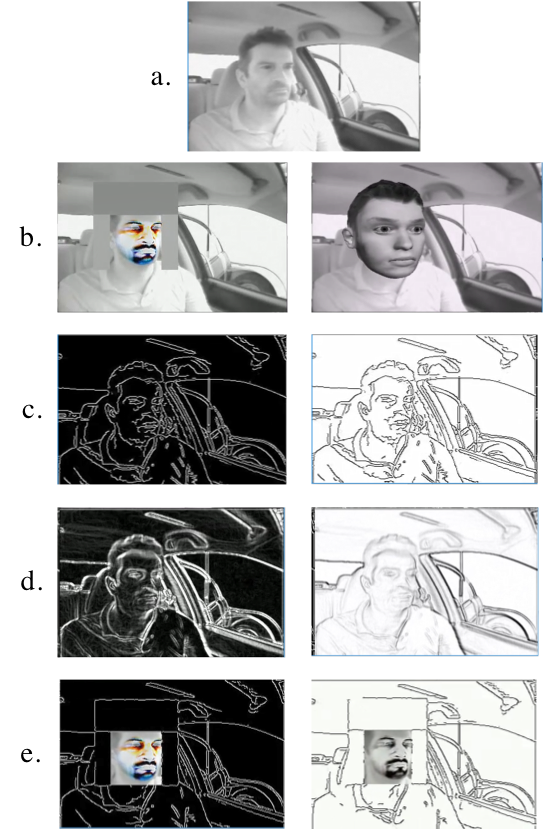}
      \caption{Mask conditions. a. Unmasked; b. ({\it left}) FAT mask, ({\it right}) DMask; c. ({\it left}) Canny, ({\it right}) Canny-inverted; d. ({\it left}) Scharr, ({\it right}) Scharr-inverted; e. ({\it left}) OverMask, ({\it right}) OverMask-inverted}
      \label{figurelabel}
   \end{figure}
 
\subsubsection{Procedure}
The experiment was composed of a study phase and a recognition phase. During the study phase, subjects studied one of two Study Lists (A and B). Each list contained high resolution, still images of half of the drivers. 
This Study List variable was used as a counterbalancing measure to assure that across all trials
in the experiment, each face would serve as a learned face and a test face an equal number of times. 
Given the odd number of images, Study List A contained 20 images, and Study List B contained 21 images. The driver images in each Study List were presented in random order and were shown 4 times during the study phase. Images were presented in the center of the screen on a black background for 4 seconds with a 2-second delay between each image presentation. No responses were required from subjects during the training. Subjects were instructed to study the pictures and were made aware that they would be asked to recognize the faces in the second phase of the experiment.  

The subsequent recognition phase was conducted immediately, during the same session, and with no significant time lapse. Subjects were assigned randomly to one of 9 mask conditions corresponding to the 8 masked and one unmasked condition. This latter served as a baseline for general recognition success on the challenging task of recognizing a person who is driving in a low-resolution video after seeing a high-resolution still image of that person in the study phase. Subjects were tested on the full set of 41 driver videos, with 20(21) familiar (i.e., studied) faces and 21(20) unfamiliar trials, depending on study list. Each subject was tested with stimuli from one of the 9 mask conditions, and the conditions were counterbalanced such that across all subjects, all mask conditions were equally represented. All videos were presented in the center of the screen on a black background. 

On each recognition trial, subjects were required to respond as to whether they recognized the driver in the video from the sequence of still images they had studied previously. Subjects indicated their response by clicking a ``YES" option on the screen for a familiar identity, or by clicking ``NEXT VIDEO" for an unfamiliar identity. Subjects could replay the videos as many times as they wished by pressing the SPACE button on the keyboard.

\subsubsection{Testing Apparatus}
The experiment was conducted using Psychophysics Toolbox \cite{Brainard1997, Pelli1997, Kleiner2007} in the MATLAB (v. R2014b) environment \cite{MATLAB:2014b}, thereby ensuring accurate stimuli presentation times. Stimuli were presented on 21.5 inch iMac systems. 

\subsubsection{Participants}
A total of 160 (125 female) undergraduate student volunteers (ages 18-47) from the University of Texas at Dallas participated in the study in exchange for research credit. 

\subsection{Results}
As in \cite{Baragchizadeh2017}, Signal Detection Theory (SDT) was employed to measure human accuracy and decision making. The SDT model was formulated for each test trial as follows. Images presented in the recognition test were either seen previously (familiar) or were not seen previously (unfamiliar). A {\it hit} was defined as a previously seen identity that was judged correctly to be familiar. A false alarm was an identity not previously seen that was incorrectly judged to be familiar.    

\subsubsection{Accuracy}
Accuracy was assessed as {\it d'}, which was calculated using the proportion of hits, $p(hit)$, and the proportion of false alarms, $p(false \ alarm)$:
 
$$d' = z(p(hit)) - z(p(false \ alarm)),$$
where the {\it z}(.) is the z-score.

Due to the non-linearity of {\it d'} for hit rates close to ceiling and false alarm rates close to floor performance, we used the Macmillan and Creelman \cite{Macmillan1991} correction for $p(hit)$ = 1 and $p(false \ alarm)$ = 0. For those rates, the correction was applied as follows. If $p(hit) == 1$, then substitute $p(hit) = 1-1/(2N_{h})$, where $N_{h}$ is the number of trials in which a familiar identity was presented. Likewise, if $p(false \ alarm) == 0$, then substitute $p(false \ alarm) = 1/(2N_{fa})$, where $N_{fa}$ is the number of trials in which an unfamiliar identity was presented.

Table 2 and Fig. 3 show the average {\it d'} and standard error of the mean (SEM) for each mask condition. These values suggest that the faces in the Unmasked condition were recognized well, and that recognition of faces in all of the masked conditions was impaired substantially.  We examine the pattern of performance across conditions more formally in {\it Section B3}.


\begin{table}[h]
\caption{Recognition Performance: Average {\it d'} in ascending order}
\label{table_example}
\begin{center}
\scalebox{1.5}{
\begin{tabular}{|l||c||l|}
\hline
 Mask Condition & Average {\it d'} & SEM\footnotemark\\
\hline
DMASK & 0.16 & 0.11\\
\hline
Canny & 0.17 & 0.1\\
\hline
OverMask-inverted & 0.23 & 0.12\\
\hline
Scharr & 0.29 & 0.09\\
\hline
FAT & 0.32 & 0.11\\
\hline
Canny-inverted & 0.32 & 0.09\\
\hline
Scharr-inverted & 0.5 & 0.11\\
\hline
OverMask & 0.59 & 0.1\\
\hline
Unmasked & 1.11 & 0.09\\
\hline
\end{tabular}}
\end{center}
\end{table}
\footnotetext{SEM is the standard error of the mean.}

   \begin{figure}[h]
      \centering
      \includegraphics[width=0.54\textwidth]{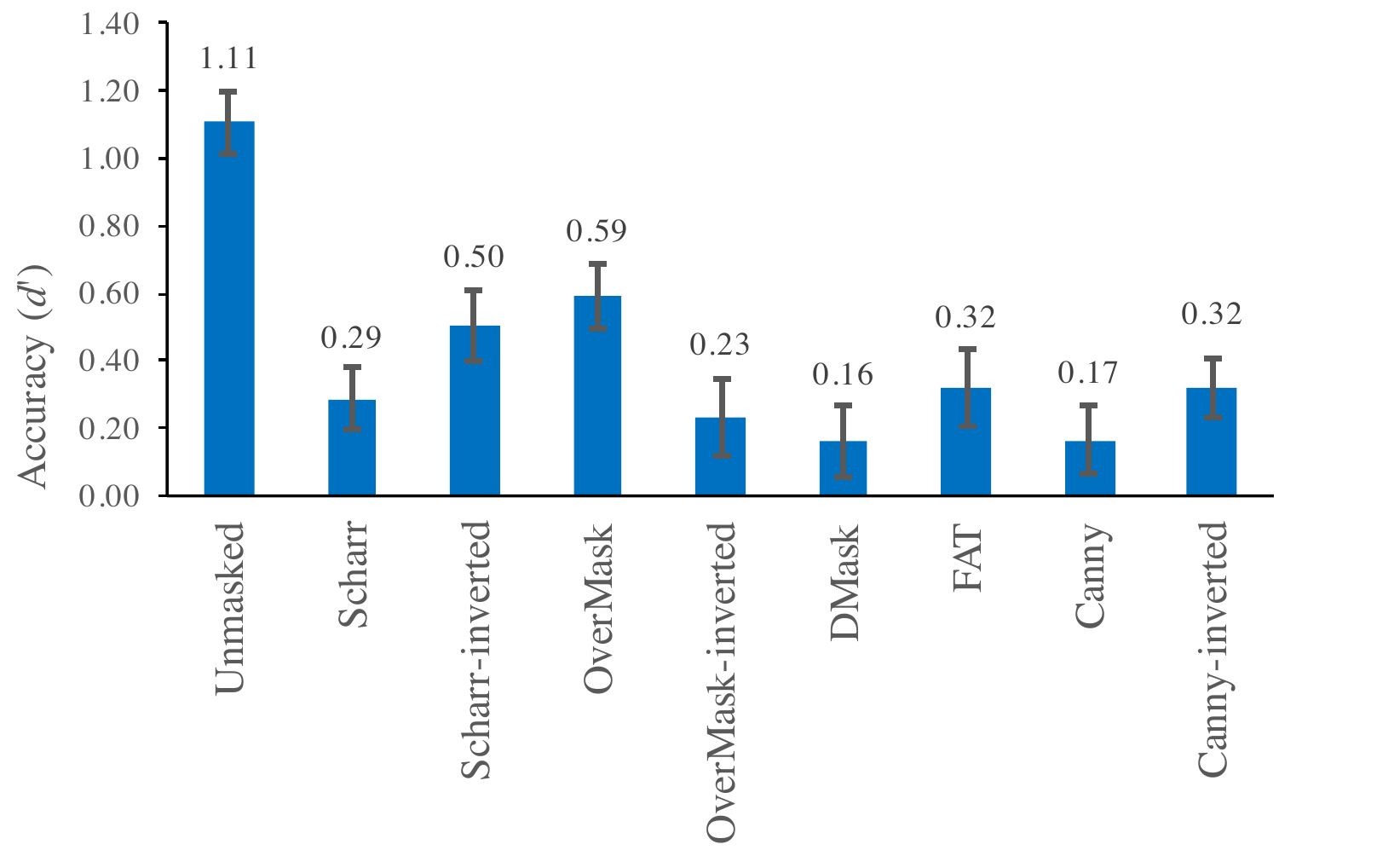}
      \caption{Average {\it d'} by Mask Condition. Error bars are SEM.}
      \label{figurelabel}
   \end{figure}

\subsubsection{Response Bias}
Response bias was measured using the signal detection criterion measure, $C$. $C$ measures a subject's propensity to make a familiar versus unfamiliar response when they are uncertain. The criterion was computed as:

$$C = -0.5 \times [z(p(hit)) + z(p(false \ alarm))]$$ 
  
This score is close to zero when subjects show no response bias; it is positive when familiar
judgments are made conservatively; and it is negative when the same judgments are made liberally.
Table 3 and Fig. 4 report the average {\it C} for all conditions. These values suggest that subjects made all of their recognition judgments conservatively, even when the test stimulus was not masked.  We examine the pattern of criteria across conditions more formally in {\it Section B3}.

\begin{table}[h]
\caption{Response Bias: Average {\it C}}
\label{table_example}
\begin{center}
\scalebox{1.5}{
\begin{tabular}{|l||c||l|}
\hline
Mask Condition & Average {\it C} & SEM\footnotemark[2]\\
\hline
Unmasked & 0.44 & 0.10\\
\hline
Scharr & 0.41 & 0.09\\
\hline
Scharr-inverted & 0.32 & 0.08\\
\hline
OverMask & 0.27 & 0.09\\
\hline
OverMask-inverted & 0.44 & 0.10\\
\hline
DMask & 0.74 & 0.19\\
\hline
FAT & 0.36 & 0.10\\
\hline
Canny & 0.29 & 0.12\\
\hline
Canny-inverted & 0.18 & 0.14\\
\hline
\end{tabular}}
\end{center}
\end{table}

   \begin{figure}[t]
      \centering
      \includegraphics[width=.54\textwidth]{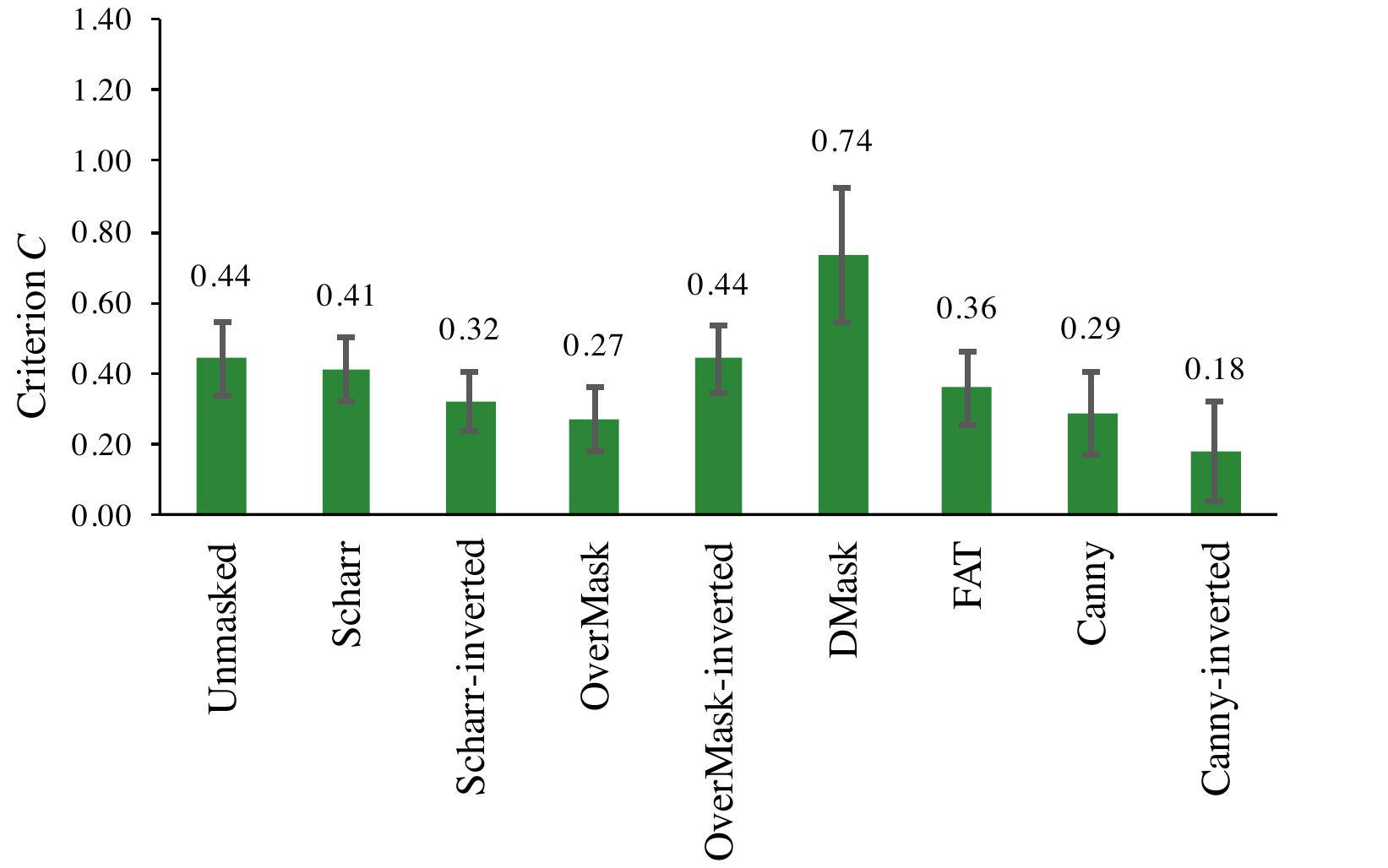}
      \caption{Average {\it C} by Mask Condition. Error bars are SEM.}
      \label{figurelabel}
   \end{figure}

\subsubsection{ANOVAs and SEM comparisons}
A two-factor Analysis of Variance (ANOVA) was conducted on the accuracy ({\it d'}) data, using Mask Condition and Study List as independent variables. There was no effect of, or interaction with, the Study List variable, and so we do not consider this variable further. The resulting one-factor model yielded a main effect of Mask Condition on {\it d'}, $F(8,142) = 9.04, p < 0.001$. 

In comparing among the conditions, the SEM values indicate that the {\it d'} accuracy was significantly higher in the Unmasked condition than the masked conditions. This indicates that all masks effectively impaired identification. Among the masked conditions, the OverMask and  Scharr-inverted conditions produced significantly higher {\it d'}s than the other masks indicating that they impaired identification less than the other masks. The Scharr, OverMask-inverted, DMask, FAT, Canny, and Canny-inverted conditions proved equally effective at impairing identification (see Figure 3). 

An analogous ANOVA was conducted for the response bias $C$. The resulting one-factor model yielded no significant main effects of Mask Condition on $C$,  $F(8,142) = 1.79, p = 0.08$ (see Figure 4).  This indicates that response bias did not differ across mask conditions
(including the no-mask condition).

Hits and false alarm rates are shown in Fig. 5.  ANOVAs were calculated separately on hit and false alarm rates. For hit rates, there was a significant main effect of Mask Condition, $F(8,142) = 3.12, p = 0.003$, with the Unmasked, Scharr-inverted, OverMask, and Canny-inverted conditions yielding the best hit rates. These conditions did not differ statistically from each other. 

Notably, there was no hit rate advantage for the Unmasked condition over the masked conditions, suggesting that the accuracy advantage seen for the Unmasked condition, as measured by {\it d'}, was due to a lower false alarm rate for Unmasked videos. This explanation was formally verified with an ANOVA on false alarm rate, which revealed a significant main effect of Mask Condition, $F(8,142) = 2.48, p = 0.015$, and a false alarm advantage for the Unmasked videos over the masked videos.

   \begin{figure}[h]
      \centering
      \includegraphics[width=.54\textwidth]{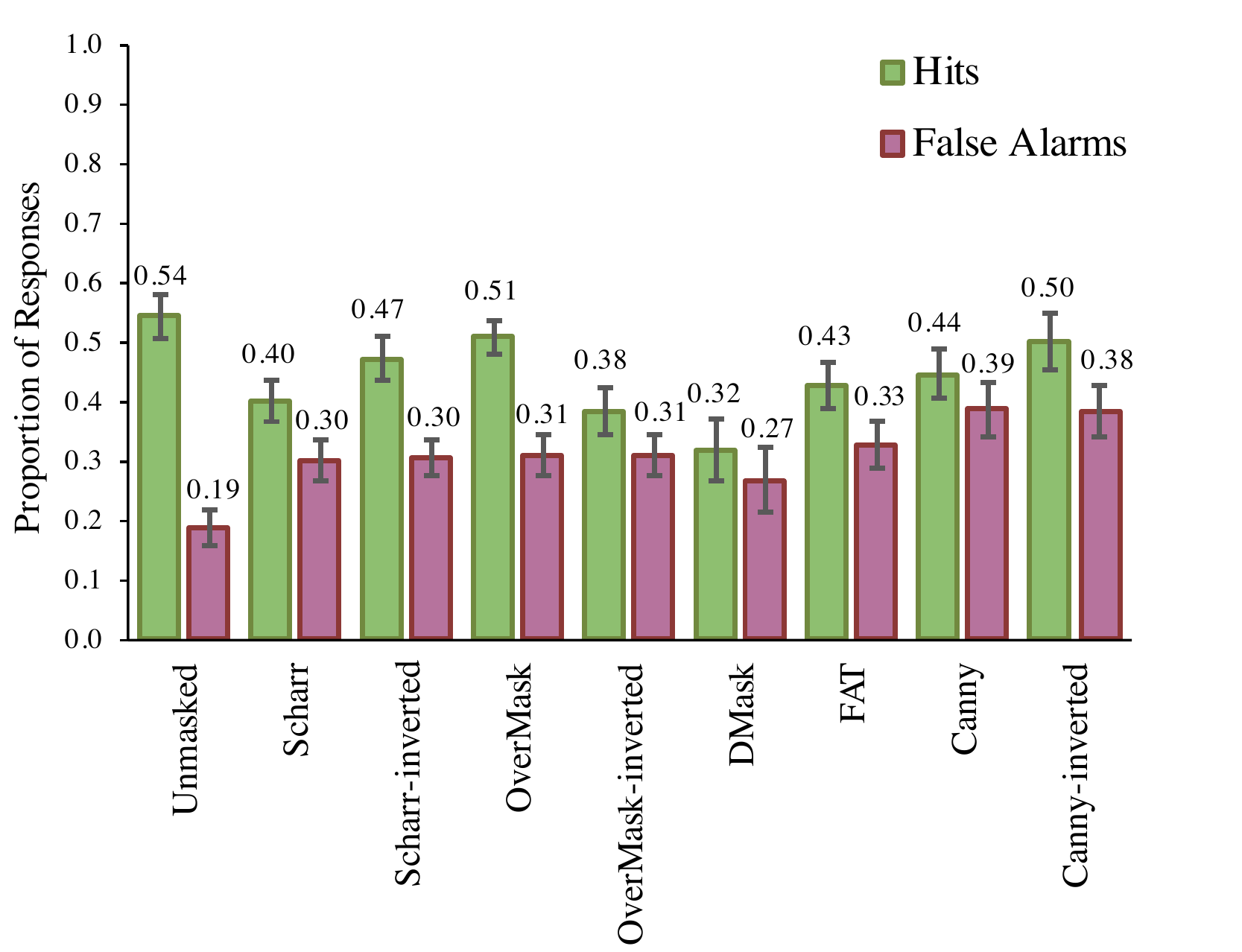}
      \caption{Hit and false alarm rates by Mask Condition. Error bars are SEM.}
      \label{figurelabel}
   \end{figure}
  
 \subsubsection{Summary: Human Recognition of Masked Faces}

The results indicate that all masks effectively impaired identification. Among the masked conditions, the OverMask and Scharr-inverted conditions were the least effective at impairing identification. The remaining masks (Scharr, OverMask-nverted, DMask, FAT, Canny, and Canny-inverted) all proved equally effective. Notably, participants used conservative criteria in all cases, indicating limited confidence in their recognition judgments.

\section{EXPERIMENT II: Evaluating the Effectiveness of Automated Identity Masking (AIM) Methods with A Deep Convolutional Neural Network (CNN)}

Next, we tested the effectiveness of identity-masking methods against identification by a state-of-the-art CNN algorithm. The goal was to determine whether the network could identify faces in the masked videos in two separate simulations. Note that for the CNN, the task was analogous to a perceptual matching task, whereby the algorithm compares a masked and unmasked face to determine if the two videos show the same person.

The following simulations were conducted: 
 
\begin{enumerate} [(i)]
\item {\it High-Resolution Images vs. Masked and Unmasked Video Templates}: This simulation involved comparing identities in high-resolution still images with those in low-resolution, masked and unmasked video frame templates. The differences in resolution, pose, and illumination between the still images and the video frames presented an extra challenge to an already complex task. To give the CNN the best chance, we created video-identity templates by averaging the video frames of each of the videos (see Methods). 

\item {\it Unmasked Video Templates vs. Masked Video Templates}: To more closely address the challenges that the resolution, pose, and illumination differences presented for the network, this simulation compared the low-resolution, unmasked videos to the low-resolution, masked videos. The masked and unmasked videos were identical frame-by-frame, except for the mask being applied to the face in the masked videos. This keeps constant factors such as perspective, illumination, trajectory of motion, etc. Although those factors might be represented by the network, they cannot provide information useful for face identification, and so this simulation provides a look at the importance of matching image conditions in network accuracy. 
 \end{enumerate}

\subsection{Methods}
\subsubsection{Dataset}
Similar to Experiment I, we used the same driver videos and still high-resolution images. We also extracted still frames from the video epochs. For each video, 76 to 77 frames were extracted.

\subsubsection{Deep Convolutional Neural Network}
The  network \cite{ranjan2018crystal} used the publicly available Face-ResNet architecture \cite{he2016} with the Universe dataset for training. Instead of using the regular Softmax Loss, by introducing the Crystal Loss (L2 Softmax) function during training, the network algorithm restricts the features to lie on a hypersphere with a fixed radius (\cite{ranjan2018crystal}).
The network consists of 101 layers with the scale factor $\alpha$ of 50. The dimension of the penultimate layer is 512 which forms the feature descriptor of an image.

\subsubsection{Procedure}
The high-resolution images and video frames were submitted to the network {\cite{ranjan2018crystal}}. The output for each image/frame was a vector of 512 features exhibiting the CNN's representation of the image in a numerical form. For each identity within a given Mask Condition, feature vectors were averaged across one video to create a template of a driver video. The cosine distance between these feature templates was used as an indicator of similarity between the templates of the drivers in masked and Unmasked conditions.


\begin{table*}[h]
\caption{Area Under the receiver operating Characteristic (AUC) scores for the high-resolution images, unmasked video templates, and masked video templates.}
\label{table_example}
\begin{center}
\scalebox{1}{
\begin{tabular}{ccccccccccc}
\hline\hline
&High-Res Image & Unmasked & FAT & DMASK & Over & Over-inverted & Canny & Canny-inverted & Scharr & Scharr-inverted\\
\hline
High-Res Image & - & 0.99 & 0.85 & 0.49 & 0.83 & 0.84 & 0.5 & 0.63 & 0.52 & 0.79\\
\hline
Unmasked & 0.99 & - & 0.99 & 0.51 & 0.99 & 0.99 & 0.56 & 0.70 & 0.54 & 0.92\\
\hline
\end{tabular}}
\end{center}
\end{table*}

\subsection{Results}
For each driver, we computed the cosine similarity scores between the feature descriptors in the high-resolution images and the averaged feature templates of each Mask Condition. Cosine similarity scores between the feature descriptors in the high-resolution images and the averaged feature templates of each of the 9 Mask Conditions were used to indicate the likelihood that the stimuli showed the same person. Analogously, cosine similarity scores between the averaged feature templates for the Unmasked and each of the 8 masked conditions indicated how similar a face in a low-resolution unmasked template was compared to its masked version. 

Face recognition accuracy was measured using the Area Under the receiver operating Characteristic (AUC). This measure indicates how capable the algorithm was at distinguishing identities. An AUC of 0.5 indicates chance performance, an AUC of 1.0 indicates perfect performance, and an AUC of 0.0 indicates that all judgments were incorrect. The AUC scores are shown in Table 4 and the corresponding Receiver Operating Characteristic (ROC) curves are plotted in Fig. 6 and Fig. 7.

   \begin{figure}[h]
      \centering
      \includegraphics[width=0.495\textwidth]{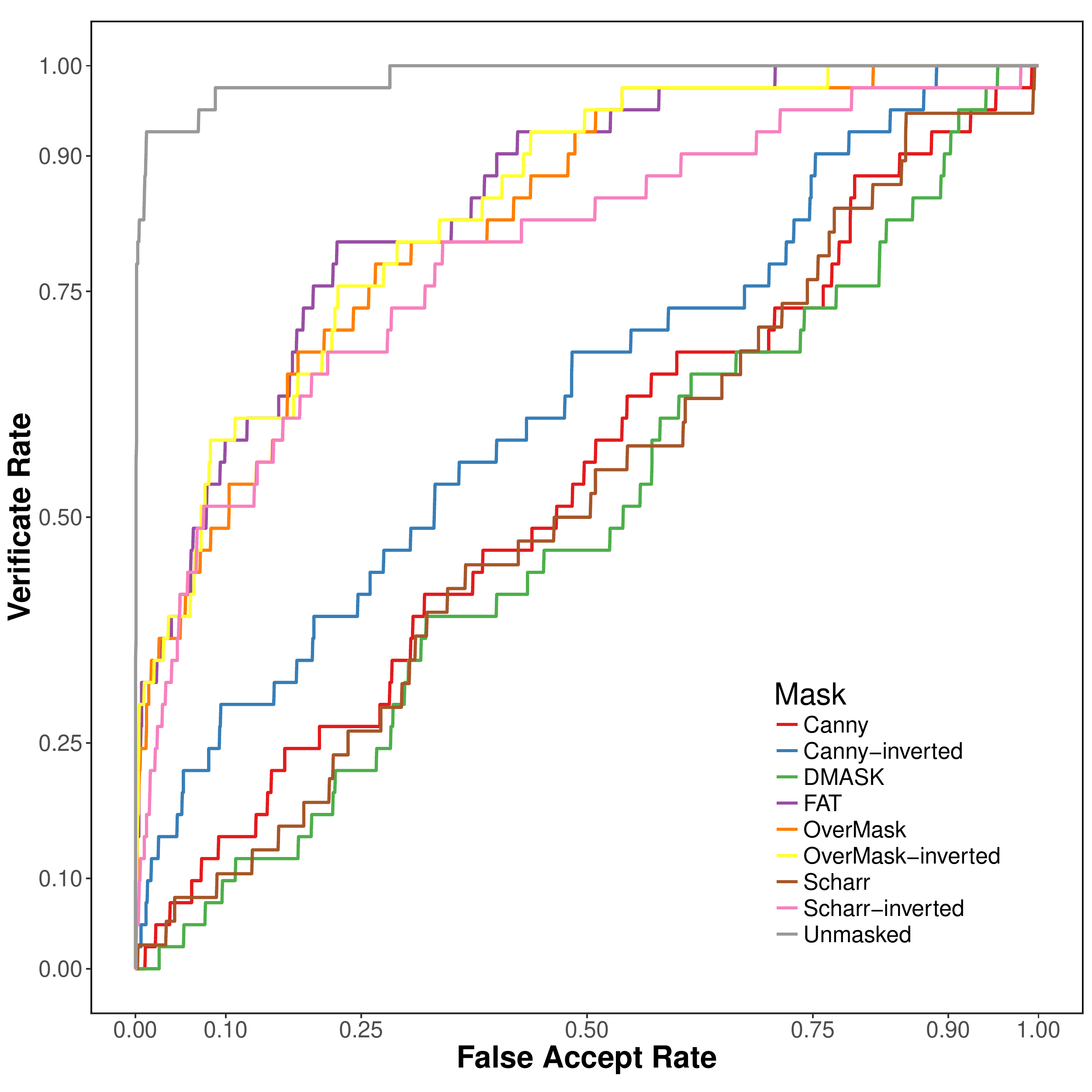}
      \label{figurelabel}
      \caption{Receiver Operating Characteristic (ROC) curves for the high-resolution images vs. the video templates from each of the 9 Mask Conditions.}
   \end{figure}
   
\subsubsection{High-Resolution Images vs. Masked and Unmasked Video Templates}
First, we measured the CNN's assessment of similarity between the high-resolution images and their corresponding video templates. The results are shown in Fig. 6 and Table IV. Performance accuracy of the network was the highest when matching the identities in high-resolution images with their unmasked video counterparts (AUC = 0.99). Network accuracy dropped when comparing high-resolution images to the FAT, Over, Over-inverted, and Scharr-inverted video templates (AUC = 0.8). Performance was at chance when the network was comparing the high-resolution images to the DMASK, Scharr, and Canny video templates (AUC = 0.5). As noted, the Over and Over-inverted masks were created by overlaying the FAT mask with the Canny and Canny-inverted masks. Because the CNN considers only the face, it was not surprising that network performance was comparable for these three mask conditions. The results are similar to those seen in the behavioral task: performance was best matching high-resolution images to unmasked videos/templates, and it suffered in all other comparison conditions. In particular, both the humans and the algorithm were at their worst when identifying drivers in the DMASK and Canny conditions.

   \begin{figure}[h]
      \centering
      \includegraphics[width=0.495\textwidth]{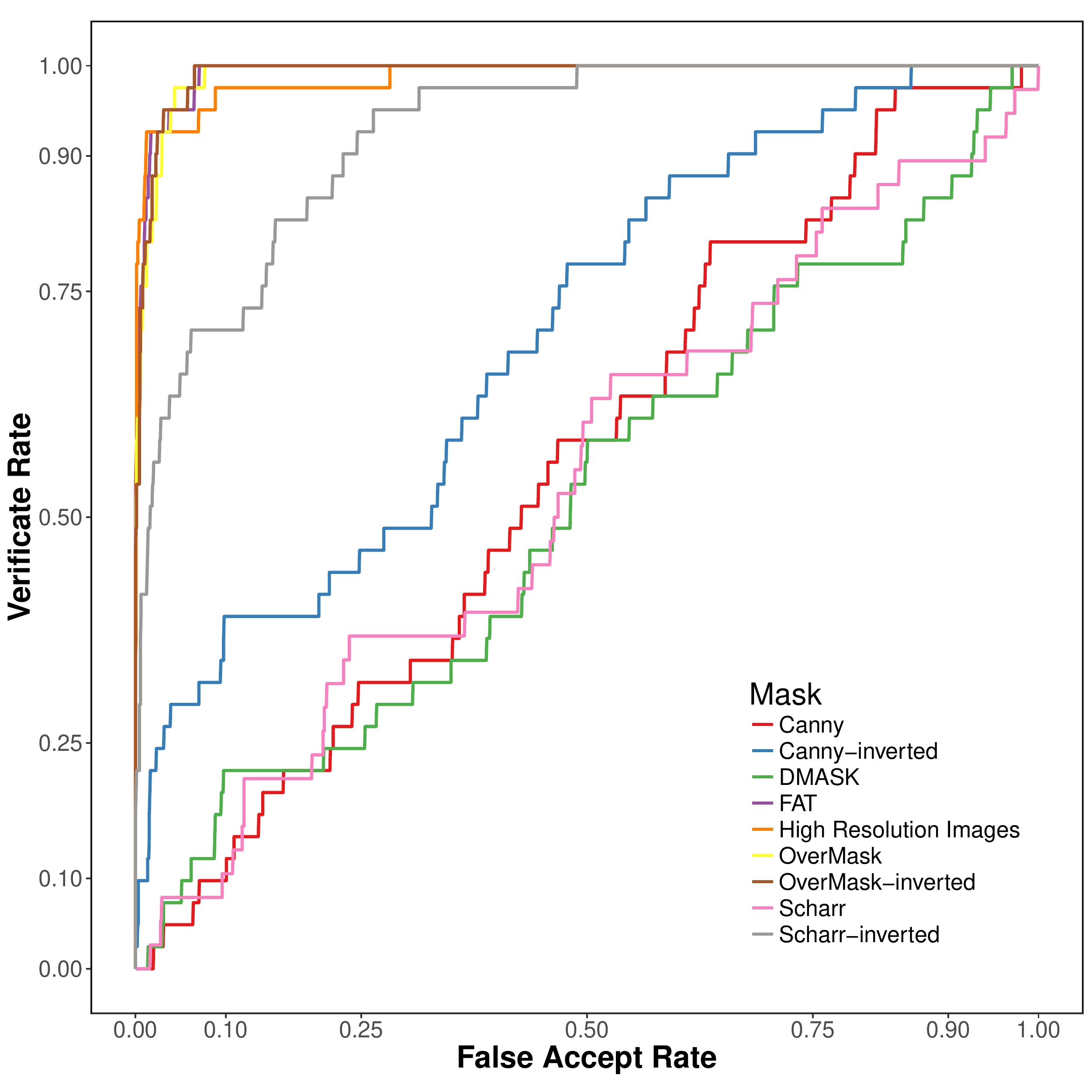}
      \label{figurelabel}
      \caption{Receiver Operating Characteristic (ROC) curves for the each of the unmasked video templates and their masked counterparts.}
   \end{figure}

\subsubsection{Unmasked Video Templates vs. Masked Video Templates}
Next, we measured the CNN's assessment of similarity between the unmasked video templates and their masked video template counterparts. The resulting ROC curves are shown in Fig. 7. Here, network performance mirrors the previous task of comparing high-resolution images to masked video templates.

\section*{Results Summary} 
In Experiment I, face recognition accuracy for human observers was impaired for all conditions in which the face was masked, relative to the Unmasked condition. The Scharr-inverted and OverMask conditions were the least effective masks. The remaining masks (Scharr, OverMask-Inverted, DMask, FAT mask, Canny, and Canny-inverted) proved highly effective in impairing recognition. Dissecting recognition into hit and false alarm rates, it is clear that recognition of the unmasked faces is highly accurate due to both a high hit rate and a low false alarm rate. Response bias was uniformly conservative and did not vary as a function of Mask Condition. 

In Experiment II, the performance accuracy of the network for matching identities in the high-resolution images to those in the unmasked video templates was at ceiling (AUC = 0.99). This validates the efficiency of the CNN despite the challenging nature of the task and the variations in illumination, pose, resolution, and compression between the high-resolution images and the video templates. 
The ordering of the performance for the masks was roughly comparable in both simulations (i.e., comparing between high-resolution and video, and comparing between unmasked video templates and masked video templates). Specifically, the best performance was found for the FAT, Over-inverted, Over, and Scharr-inverted masked video templates, followed by the Canny-inverted condition. The lowest accuracy was seen in the Canny, Scharr, and the DMask conditions. The low accuracy observed in the DMask condition was not surprising, as this mask is a generic avatar face mapped on the face of the driver. However, the differences between the Scharr/Scharr-inverted, and Canny/Canny-inverted conditions are remarkable given the limited difference of inversion between these condition pairs. Whereas performance accuracy in the Canny condition was at chance, accuracy was substantially
higher in the Canny-inverted condition (AUC = 0.7). This trend is also evident between the Scharr and Scharr-inverted conditions - a simple color inversion yields higher accuracy.

In summary, the results for both human perceivers and the CNN indicated that, compared to other mask conditions, the DMask and Canny conditions were most successful at impairing the network's identification performance.

\section*{Discussion and future work} 

Our aims were: a.) to compare a variety of identity-masking algorithms using human perception as the ultimate standard of recognition performance; and b.) to test the effectiveness of identity masking with a state-of-the-art CNN for the same task. Our results show that the effectiveness of identity masking varies as a function of mask condition for both human observers and the CNN. Given the results of both humans and machines, we speculate that the use of a one-size-fits-all masking algorithm is perhaps not an ideal approach to identity masking. In addition, the various challenges posed in video stimuli (e.g., lighting, shadows, movement, resolution, etc.) might be best overcome using a combinatorial approach.

We also addressed questions that remained following the work of \cite{Baragchizadeh2017}. First, we eliminated the possibility that recognition could be based on information other than the face. Specifically, we eliminated hair and ears as possible cues to identity in the training images shown to human observers. 
Second, we tested subjects only on stimuli in which the mask was successful (i.e., eliminating short epochs in the video where the face tracking algorithm failed due to abrupt movements.) In the previous work \cite{Baragchizadeh2017}, natural movements of the drivers sometimes caused problems in facial tracking, causing the mask to fail to adhere to the face. The current research uses motion-limited video epochs to eliminate this facial-tracking failure. Additionally, we used a combinatorial masking method, the Overmask method, which combines filtering and avatar approaches. Although weaker at masking, filtering methods do not suffer from tracking failures and so are not limited by abrupt motions. Conversely, avatar-based methods are ineffective when the mask misses the face center, but are highly effective when the mask is properly placed. We note that combinatorial methods such as Overmask may be a good option in less-controlled stimulus sets, when abrupt motions are likely.
 
Third, participants' confidence ($C$) in their identifications was always above zero, indicating conservative response tendencies for every mask condition. Human face identifications made with weak confidence are not usually accorded strong credibility. Combined, the present effects support the practice of using data license agreements that specifically prohibit attempts to re-identify the drivers, and they make a strong case for the benefits of masking these data sets to improve data access for transportation studies.

Fourth, we expanded the number of identity-masking algorithms, comparing 8 algorithms based on diverse methods. Although all of the masks impaired identification substantially over the unmasked face, none of the masks completely eliminated identifiability. In addition, differences in mask effectiveness (as measured via {\it d'}) were observed between the algorithms tested, including algorithms that were simply direct inversions of other algorithms. These differences highlight the relative strength of different masks for different video challenges. 

Fifth, we added results from a subset of masking algorithms tested using a state-of-the-art deep convolutional neural network. Although no formal comparison between CNN and human observer performance was made here, we note that the overall behavioral results were corroborated by the performance of the CNN in the second experiment.

Given our observations of CNN performance, future work could evaluate the effectiveness of using CNNs to predict which mask algorithms are better suited for certain stimuli. This kind of stimulus-specific identity-masking could be the next step in preserving privacy in an age of continuous video surveillance.

\section*{ACKNOWLEDGMENTS}
This work was supported through collaboration with Oak Ridge National Laboratory and the Federal Highway Administration under the Exploratory Advanced Research Program (Contracting Officer's Representative: Lincoln Cobb). The human experiment and analysis was subcontracted to the University of Texas at Dallas from Oak Ridge National Laboratory. The CNN feature extraction was carried out at University of Maryland
by C. C., who was  supported by the Intelligence Advanced Research Projects
	Activity (IARPA). The UMD part of the research is based upon work supported by the Office
	of the Director of National Intelligence (ODNI), Intelligence Advanced
	Research Projects Activity (IARPA), via IARPA R\&D Contract No.
	2014-14071600012. The views and conclusions contained herein are those of
	the authors and should not be interpreted as necessarily representing the
	official policies or endorsements, either expressed or implied, of the ODNI,
	IARPA, or the U.S. Government. The U.S. Government is authorized to
	reproduce and distribute reprints for Governmental purposes notwithstanding
	any copyright annotation thereon.


\addtolength{\textheight}{-1 cm}   

\bibliographystyle{IEEEtran}
\bibliography{AIM2_CNN_arXiv}

\end{document}